\documentclass{article}

\usepackage{arxiv}

\usepackage[utf8]{inputenc} % allow utf-8 input
\usepackage[T1]{fontenc}    % use 8-bit T1 fonts
\usepackage{hyperref}       % hyperlinks
\usepackage{url}            % simple URL typesetting
\usepackage{booktabs}       % professional-quality tables
\usepackage{amsfonts}       % blackboard math symbols
\usepackage{nicefrac}       % compact symbols for 1/2, etc.
\usepackage{microtype}      % microtypography
\usepackage{lipsum}
\usepackage{graphicx}
\usepackage{amsmath}
\graphicspath{ {./images/} }
\usepackage{xcolor}
\usepackage[noabbrev, capitalise]{cleveref} % no nameinlink
\usepackage{authblk}
\usepackage{soul}
\usepackage{subcaption}
\usepackage{listings}
\usepackage[flushleft]{threeparttable}
\usepackage{booktabs}
\usepackage{makecell}
\usepackage{multirow}

\hypersetup{
    colorlinks=true,
    linkcolor=blue,
    filecolor=magenta,      
    urlcolor=blue,
    }

\title{CarDreamer: Open-Source Learning Platform for World Model based Autonomous Driving}
\author[1]{% 
\bf
Dechen Gao$^{*}$}%
\author[2,4]{%
\bf	Shuangyu Cai$^{*}$}%
\author[2]{%
\bf	Hanchu Zhou$^{*}$}%
\author[2]{%
\bf	Hang Wang$^{*}$}%
\author[3]{%
\bf	Iman Soltani}
\author[2]{\bf Junshan Zhang}
\affil[1]{Department of Computer Science, University of California, Davis}
\affil[2]{Department of Electrical and Computer Engineering, University of California, Davis}
\affil[3]{Department of Mechanical and Aerospace Engineering, University of California, Davis}
\affil[4]{Institute for Interdisciplinary Information Sciences, Tsinghua University, China}
\affil[ ]{ \url{{dcgao,hczhou,whang,isoltani,jazh}@ucdavis.edu}
}
\affil[4]{\url{caisy21@mails.tsinghua.edu.cn}}
\affil[$^{*}$]{Equal contribution.}

\begin{document}
\maketitle

\begin{abstract}
To safely navigate intricate real-world scenarios, autonomous vehicles (AVs) must be able to adapt to diverse road conditions and anticipate future events. World model based reinforcement learning (RL) has emerged as a promising approach by learning and predicting the complex dynamics of various environments. Nevertheless, to the best of our knowledge,  there  does not exist an open-source platform for   training and testing such algorithms in complicated driving environments.
To fill this void, we introduce CarDreamer, the first open-source learning platform designed specifically for developing and evaluating world model based autonomous driving algorithms. It comprises a few key components, including
1) World model (WM) backbone: CarDreamer has integrated some state-of-the-art world models, which simplifies the reproduction of RL algorithms. The backbone is decoupled from the rest and communicates using the standard Gym interface, so that users can easily integrate and test their own algorithms.
2) Built-in tasks:  CarDreamer offers a comprehensive set of highly configurable driving tasks which are compatible with Gym interfaces and are equipped with empirically optimized reward functions.
3) Task development suite: CarDreamer integrates a flexible task development suite to streamline the creation of driving tasks. This suite enables easy definition of traffic flows and vehicle routes, along with automatic collection of multi-modal observation data. A visualization server allows users to trace real-time agent driving videos and performance metrics through a browser.
Furthermore, we conduct extensive experiments using built-in tasks to evaluate the performance and potential of WMs in autonomous driving.
Thanks to the richness and flexibility of CarDreamer, we also systematically study the impact of observation modality, observability, and sharing of vehicle intentions on AV safety and efficiency. All code and documents are accessible on our GitHub page \url{https://github.com/ucd-dare/CarDreamer}.

\end{abstract}

% keywords can be removed

\section{Introduction}

Autonomous vehicles (AVs) are expected to play a central role in future mobility system with many promising benefits like safety and efficiency~\cite{grigorescu2020survey}. Recent years have witnessed great achievement on the development of AVs. In the U.S. alone, millions of miles have been driven by AVs~\cite{schwall2020waymo} on public roads. However, achieving robust AVs that are capable of navigating complex and diverse real-world scenarios remains a challenging frontier~\cite{yurtsever2020survey,teng2023motion,chen2023end}. For instance, as calculated by the US Department of Transportation’s Federal Highway Administration, AVs experience a crash rate about two times more than the conventional vehicles per million miles traveled~\cite{crashrate}.

The reliability of AVs directly hinges upon the generalization capability of autonomous systems in unforeseen scenarios. World model (WM), which excels in generalization, offers a promising solution with its ability to learn the complex dynamics of environments and anticipate future scenarios. In particular, WMs learn a compact latent representation that encodes the key elements and dynamics of the environment. This learned representation facilitates strong generalization, enabling the WM to make predictions in scenarios beyond its training samples. In particular, our WM-based learning platform incorporates building blocks that mimic human-like perception and decision-making, such as vision model and memory model~\cite{ha2018world,hafner2019dream}. Indeed, humans excel at handling rare or unseen events with proper actions thanks to human's ``internal world model''~\cite{hu2022model}. By emulating cognitive processes akin to human intelligence, WM based reinforcement learning (RL) has demonstrated state-of-the-art performance in domains such as Atari games and Minecraft~\cite{hafner2023mastering}. However, WMs' application on autonomous driving remains an exciting open field~\cite{chen2023end}, partially due to the lack of easy-to-use platforms to train and test such RL algorithms. The endeavor on developing a learning platform for WM-based autonomous driving can be extremely beneficial for the research in this domain. 

Thus motivated, we introduce CarDreamer, the first open-source learning platform designed specifically for WM based autonomous driving. CarDreamer aims to facilitate the rapid development and evaluation of algorithms, enabling users to test their algorithms on provided tasks or quickly implement customized tasks through a comprehensive development suite. CarDreamer's  key contributions include:

\begin{enumerate}

    \item \textit{Integrated WM algorithms for reproduction.} CarDreamer has integrated state-of-the-art WMs, including DreamerV2, DreamerV3, and Planning2Explore, significantly reducing the time required to reproduce the performance of existing algorithms. These algorithms are decoupled from the rest of CarDreamer and communicates through the unified Gym interface. This enables streamlined integration and testing of new algorithms without additional adaptation efforts as long as they support Gym interface.

    \item \textit{Highly configurable built-in tasks with optimized rewards.} CarDreamer provides a comprehensive set of driving tasks, such as lane changing and overtaking. These tasks allow extensive customization in terms of difficulty, observability, observation modalities, and communication of vehicle intentions. They expose the same Gym interface for convenient use and the reward functions are meticulously designed to optimize training efficiency.
    
    \item \textit{Task Development Suite and Visualization Server.} This suite not only simplifies the creation of customized driving tasks via API-driven traffic spawning and control,  but also includes a modular observer for easy multi-modal data collection and configuration. A visualization server enables the real-time display of agent driving videos and statistics on a web browser, which accelerates reward engineering and algorithm development by providing immediate performance insights.

\end{enumerate}

In addition to developing the CarDreamer platform, we present comprehensive experiments that evaluate the overall performance and potential of WMs in autonomous driving. We highlight its predictive accuracy on multi-modal observation inputs. Furthermore, the comparison of different levels of observability and intention sharing demonstrates that communication can clearly enhance both traffic safety and efficiency. To the best of our knowledge, these results represent the first experimental manifestation of WMs’ efficacy in autonomous driving tasks with communication of vehicle intentions.

% This paper details the main structure of CarDreamer and illustrates its practical application through a series of WM based experiments in autonomous driving.

% Conventional  autonomous vehicles approaches usually consist of perception, prediction, planning, and control modules~\cite{jia2023adriver}. The perception module is leveraged to detect surrounding vehicles, pedestrians, obstacles, which are later after used by a prediction module to forecast their future trajectories.  autonomous vehicles then plan future positions and actuate low-level control signals given the detection and prediction. Though autonomous driving has witnessed great success in recent years,  autonomous vehicles can still struggle with driving scenarios that human address with ease...

% To that end, world models have emerged as a promising approach, since the agent policy can be trained in world models' latent ``dreams" to improve sample efficiency and generalizability. However.

\section{Related Work}
\label{sec:related}

\textbf{World Models in Reinforcement Learning.} RL usually suffers from high sample complexity~\cite{yu2018towards}, which significantly hinders it from being practical especially for tasks where interacting with environment can be costly and time-consuming. To remedy for the issue, model-based RL  has been studied to leverage a world model that explicitly learns environment dynamics to ``imagine'' future trajectories, allowing agents to interact with the world model instead of the actual environment~\cite{ha2018world}. As the high-dimensional observations that evolve on intricate dynamics can be intractable, prior works typically learn dynamics on a latent space. Typical designs involve modeling dynamics with a Recurrent State-Space Model (RSSM)~\cite{hafner2019learning}. Dreamer provides a series of works that leverage RSSM~\cite{hafner2019learning, hafner2020mastering, hafner2023mastering} and train agents in world model's imagination,  and has demonstrated promising sample efficiency and generalization ability on conventional RL benchmarks. ISO-Dream~\cite{pan2022iso} isolates controllable and non-controllable sources to the changes of dynamics such that agents can differentiate changes that are independent and dependent to their actions. Planning2Explore~\cite{sekar2020planning} promotes exploration by directing it towards states of higher uncertainty, facilitating the learning of more robust dynamics and enabling quick adaptation to new tasks in a zero or few-shot manner. LEXA \cite{mendonca2021discovering} utilizes a learned world model to train separate explorer and achiever policies for forward-looking exploration and goal achievement.

\textbf{World Models for Autonomous Driving.} In the field of autonomous driving, there has been mainly two approaches for studying world models~\cite{guan2024world, zhu2024sora}, 1) leveraging world models as neural driving simulators to synthesize realistic driving videos, and 2) utilize world models in simulation to train and evaluate agent policies. In the first approach, GAIA-1~\cite{hu2023gaia} utilizes a world model to generate realistic driving scenarios given videos, texts, and actions as inputs. DriveDreamer~\cite{wang2023drivedreamer} generates driving scenarios along with the actions given prior information such as high-definition maps and 3D bounding boxes. ADriver-1~\cite{jia2023adriver} eliminates the need of extensive prior information and is able to achieve ``infinite driving'' inside its world model by providing both scenario generation and action predictions. DriveDreamer-2~\cite{zhao2024drivedreamer} is built-upon large language models to make prompting more user-friendly to generate diverse traffic conditions for driving video generation. In contrast, in the second approach, MILE~\cite{hu2022model} conducts imitation learning based on a Dreamer-style world model. It learns from offline expert data using road map and camera inputs, and predicts the transition of future Bird-Eye Views (BEVs) as an auxiliary task. SEM2~\cite{gao2022enhance} conducts reinforcement learning with a Dreamer-style world model, and decodes camera and LiDAR representations into semantic BEV masks.  Think2Drive~\cite{li2024think2drive} trains DreameV3 with BEV inputs and tests it on 39 CARLA benchmarks. Our platform aims at facilitating this line of research, providing tailored benchmarks and tools for WM based RL algorithms in autonomous driving.

\textbf{Simulators.} Collecting data for autonomous driving in the  real world is costly and time-consuming. To this end, various simulators, such as CARLA~\cite{dosovitskiy2017carla}, SUMO~\cite{lopez2018sumo}, and Flow~\cite{wu2017flow}, have been developed. CARLA is distinguished by its realistic environmental modeling and image rendering capabilities. These simulators are generally designed for general traffic simulation rather than for RL applications. Our platform is specifically tailored for WM-based RL, offering meticulously crafted RL rewards, interfaces, and automating training data requirements.

\section{Background}

We give a brief introduction in this section to the two cornerstones that CarDreamer involves: CARLA,~\cite{dosovitskiy2017carla} a high-fidelity and flexible simulator, and Gym~\cite{brockman2016openai}, a standard interface for RL training and evaluation.

\textbf{CARLA.} CARLA is an open-source simulator that aims at simulating real-world traffic scenarios. CARLA is based on Unreal Engine which provides realistic physics and high-quality rendering. It provides digital assets including maps, buildings, vehicles, and various landmarks. It supports various sensors such as RGB camera, LiDAR, RADAR. Users can create vehicles or pedestrians and take full control of these actors. It is  a general tool, and one main challenge of its application in RL algorithms also originates from its generality: As  stated in \cref{sec:related}, obtaining the BEV in CARLA involves a cumbersome process, thus impeding its fast deployment in training RL algorithms.

\textbf{Gym.} Gym is a standard interface defined by OpenAI to formalize the interplays  between agents and environments. Two functions \lstinline{reset()} and \lstinline{step(action)} constitute the core part of this interface. The former initializes the environment to its start state. The latter takes an action input from the agents, simulates the evolution of the environment, and returns observation data, reward signals, a terminal indicator, and some extra information. In this way, an RL algorithm can be easily tested on various environments with minor adaptation as long as they both support the Gym interface. There have been extensive efforts in developing diverse Gym benchmarks, such as Atari games and DMC suites. To the best of our knowledge, in the field of WM based RL algorithms for autonomous driving in CARLA, CarDreamer is the first platform that provides diverse urban driving tasks through the Gym interface to facilitate training and evaluation.

\section{CarDreamer Architecture and Implementation}
\begin{figure*}
    \centering
    \includegraphics[width=\linewidth]{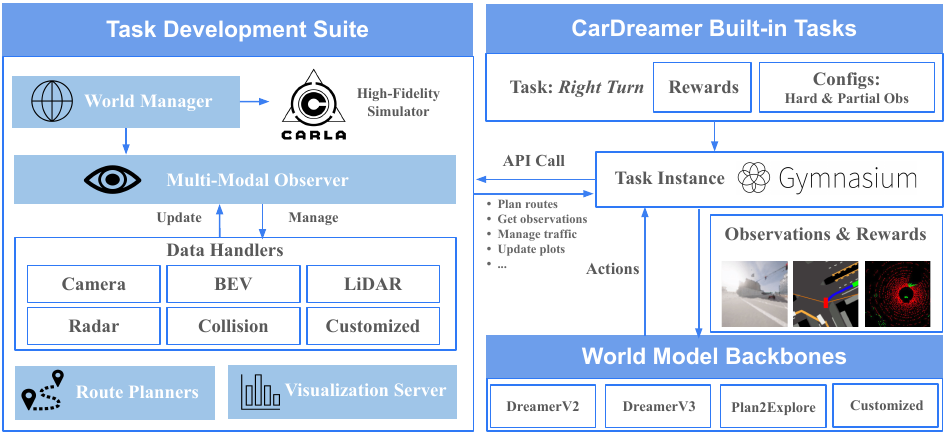}
    \caption{CarDreamer Architecture. Three key components are highlighted: Built-In Tasks, Task Development Suite, World Model Backbone.}
    \label{fig:method_architecture}
\end{figure*}

% \begin{figure}
%     \centering
%     \includegraphics[width=0.9\textwidth]{figures/workflow.JPG}
%     \caption{CarDreamer Workflow.}
%     \label{fig:method_architecture}
% \end{figure}

% The primary goal of CarDreamer is to facilitate the implementation and development of world model based methods for autonomous driving. Each module is designed to be both efficient and flexible such that they can be easily modified or adjusted to satisfy users' own research needs.

\subsection{Overview of CarDreamer}

As depicted in \Cref{fig:method_architecture}, CarDreamer comprises three principal components: \textit{world model backbone}, \textit{built-in tasks}, and \textit{task development suite}. In particular, the task development suite facilitates a variety of API functionalities, including vehicle spawning, traffic flow control, and route planning within CARLA. An observer module automates the collection of multi-modal observation data, such as sensor data and BEVs, managed by independent and customizable data handlers. This data serves dual purposes: it is utilized by the task and a training visualization server. The visualization server displays real-time driving videos and environment feedback via an HTTP server and integrates seamlessly with the world model algorithm through the gym interface. Upon receiving an action as the agent's response, the observer collects data from data handlers at the subsequent frame, thus continuing this operational cycle. We will now explore each module in detail.

\subsection{World Model Backbone}  
The World Model Backbone in CarDreamer seamlessly integrates state-of-the-art approaches such as DreamerV2~\cite{hafner2020mastering}, DreamerV3~\cite{hafner2023mastering}, and Planning2Explore~\cite{sekar2020planning}, facilitating rapid reproduction of these models. This backbone architecture is strategically designed to decouple the world model implementation from task-specific components, thereby enhancing modularity and extensibility. Communication between these components is efficiently managed through the standard Gym interface, which allows for extensive customization.

This decoupling enables users to easily adapt or replace the default world models with their own implementations, supporting rapid prototyping, benchmarking, and comparative analysis against established baselines. CarDreamer thus provides a comprehensive testbed for world model-based algorithms, fostering an ecosystem conducive to accelerated research and development within this field. The platform encourages users to explore innovative architectures, loss functions, and training strategies, all within a consistent and standardized evaluation framework characterized by diverse driving tasks and performance metrics.

\subsection{Built-In Tasks}

We have meticulously crafted a wide array of realistic tasks, ranging from simple skills such as lane following and left turning to more complex challenges like random roaming across mixed road conditions that include crossroads, roundabouts, and varying traffic flows. These tasks are highly configurable, offering numerous options that present fundamental questions in autonomous driving.

\textbf{Observability \& Intention Sharing:} Partial observability presents a significant challenge in RL, where incomplete state information can exponentially increase the complexity of the input space for the sake of encompassing  historical steps~\cite{foerster2016learning}. To address the lack of tools tailored to these challenges in autonomous driving, we initially offer three observability settings in CarDreamer: 1) Field-of-View (FOV) includes only the vehicles within the camera's FOV; 2) Shared-FOV (SFOV) enables a vehicle to communicate with and collect FOV data from other vehicles within its own FOV; 3) Full Observability (FULL) assumes complete environment and background traffic information. Furthermore, users have control over whether the vehicles share their intention, and whom the vehicles share with. These configurations aligns with the fundamental questions of ``what information to communicate'' and ``whom to communicate with''~\cite{zhu2022survey}.

\textbf{Observation Modality:} Users can configure the observation space to include various modalities, from sensor data such as RGB cameras and LiDAR to synthetic data such as BEVs. This flexibility supports the development of end-to-end models that are capable of making decisions directly from multi-modal raw sensor data~\cite{chen2023end} or planning with BEV perception~\cite{chen2020learning}.

\textbf{Difficulty:} Difficulty settings primarily affect the density of traffic, posing significant collision avoidance challenges. As safety-critical events of AVs are rare~\cite{feng2023dense}, it is inherently nontrivial to validate the robustness of AVs due to the infrequent nature of such events~\cite{kalra2016dts}. CarDreamer is specifically designed to enable a comprehensive evaluation of safety and efficiency in scenarios that mimic these infrequent but critical events.

\textit{Reward function.} Each task within CarDreamer is equipped with an `optimized' reward function, which has been experimentally shown to enable DreamerV3 to successfully navigate through waypoints within just 10,000 training steps (see \Cref{sec:results} for details). Notably, our empirical findings indicate that rewarding the agent based on its speed or incremental position changes leads to superior performance compared to rewarding absolute position. This is because when be rewarded solely for position, the agent can exploit the reward function by making a small initial movement and then remaining stationary, as any further movement risks incurring collision penalties. In practice, we do observe such sub-optimal behavior, where the learned policy converges to a local optimum to avoid collision by remaining stationary with a constant position reward. Conversely, rewarding speed encourages the agent to keep moving and accumulate rewards, mitigating the risk of premature convergence to sub-optimal policies. 

Our reward design carefully addresses crucial requirements for driving tasks, such as trajectory smoothness, which are often overlooked in conventional RL algorithms. Typically, these algorithms include an entropy term in their loss function or value estimation to encourage exploration and prevent premature convergence. However, in autonomous driving contexts, this entropy term can incentivize vehicles to follow a zigzag trajectory, as such erratic motion generates higher entropy rewards compared to smoother paths, even though both trajectories might achieve similar progress towards the goal. To counteract this effect, we introduce a penalty term specifically designed to discourage motion perpendicular to the goal direction. As a result, we have developed a reward function that effectively balances goal progression and trajectory smoothness, structured as follows

\begin{equation}
    r = \alpha v_{\text{parallel}} - \beta v_{\text{perp}} - \gamma \mathbb{I}_{\text{collision}}.
\end{equation}

Here, $v_{\text{parallel}}$ and $v_{\text{perp}}$ represent the speed parallel and perpendicular to the goal direction, respectively. $\mathbb{I}_{\text{collision}}$ is the indicator for collision. $\alpha$, $\beta$, $\gamma$ are scaling factors. For tasks like waypoint following, additional reward terms are included for reaching each waypoint.

\textit{Interface and Usage.} All built-in tasks in CarDreamer utilize a unified Gym interface, allowing simple-to-use training and testing of RL algorithms without additional adaptations. Beyond direct usage, CarDreamer supports a variety of algorithms, including those for curriculum learning, which can leverage the progression from simpler to more complex tasks; or continual learning, which aims at addressing catastrophic forgetting when learning a new task. Additionally, for imitation learning, CarDreamer simplifies the collection of observational data in the simulator. Although initially designed for WM-based RL algorithms, the gym interface enables diverse applications across various algorithmic strategies.

\subsection{Task Development Suite}
For users requiring customized tasks, CarDreamer offers a highly modular Task Development Suite. This suite is adaptable to various levels of customization to satisfy diverse user requirements.

The initial module, \textit{World Manager}, caters to basic needs such as varying driving scenarios with different maps, routes, spawning locations, or background traffic flows. The World Manager is responsible for managing `actors,' a term from CARLA~\cite{dosovitskiy2017carla}, which encompasses all entities including vehicles, pedestrians, traffic lights, and sensors. It provides API calls to spawn various actors, particularly vehicles at different locations with either a default or a customized blueprint. These vehicles can be controlled by the user or by an Autopilot, a simple rule-based autonomous driving algorithm. Upon reset, it transparently destroys and releases resources.

The second module, the \textit{Observer}, automates the collection of observation data across various modalities. While it allows users to easily access pre-defined observation modalities without manual interaction, it also supports extensive customization for data specifications. This is achieved through a series of data handlers, each delivering data for a particular modality, such as a RGB camera handler and a BEV handler. Each data handler is highly modular and independently manages the entire lifecycle of a specific type of data. Users can enhance the observer by registering a new data handler tailored to their own requirements.

The third module comprises \textit{Route Planners} that accommodate diverse needs for task routes. CarDreamer includes several planners: a random planner for exploratory roaming across the entire map, a fixed path planner that creates waypoints connecting user-defined locations, and a fixed ending planner that generates routes using the classical A* algorithm from the current position to a designated endpoint. For additional customization, a base class is available for users to develop their own planners by overriding the \lstinline{init_route()} and \lstinline{extend_route()} methods, which define the initialization and extension of routes per time step, respectively.

Additionally, the suite features a visualization server that seamlessly integrates the output from the Observer and other statistical data from environment feedback, displaying via an HTTP server. This automation facilitates rapid feedback, enhancing the process of reward engineering and algorithm development without extra coding efforts.

\section{CarDreamer Task Experiments}
\label{sec:results}

This section showcases the versatility and capabilities of CarDreamer through a comprehensive set of experiments across a wide range of settings.  We use DreamerV3 \cite{hafner2023mastering} as the model backbone. \Cref{sub:train} focuses on task training and evaluation, where we evaluate the performance of WMs in diverse driving tasks within CarDreamer. In \Cref{sub:open}, we assess the prediction accuracy of WMs to accurately imagine future states in different observation modality settings.
% \Cref{sub:obsmod} investigates the impact of different observation modality, such as BEV, LiDAR, and RGB camera, on the performance of WM based algorithms.
Furthermore, \Cref{sub:com} systematically evaluates the significant impact of observability and intention sharing on traffic safety and efficiency.

\subsection{World Model Training \& Evaluation} \label{sub:train}
\begin{figure}
    \centering
    \includegraphics[width=\textwidth]{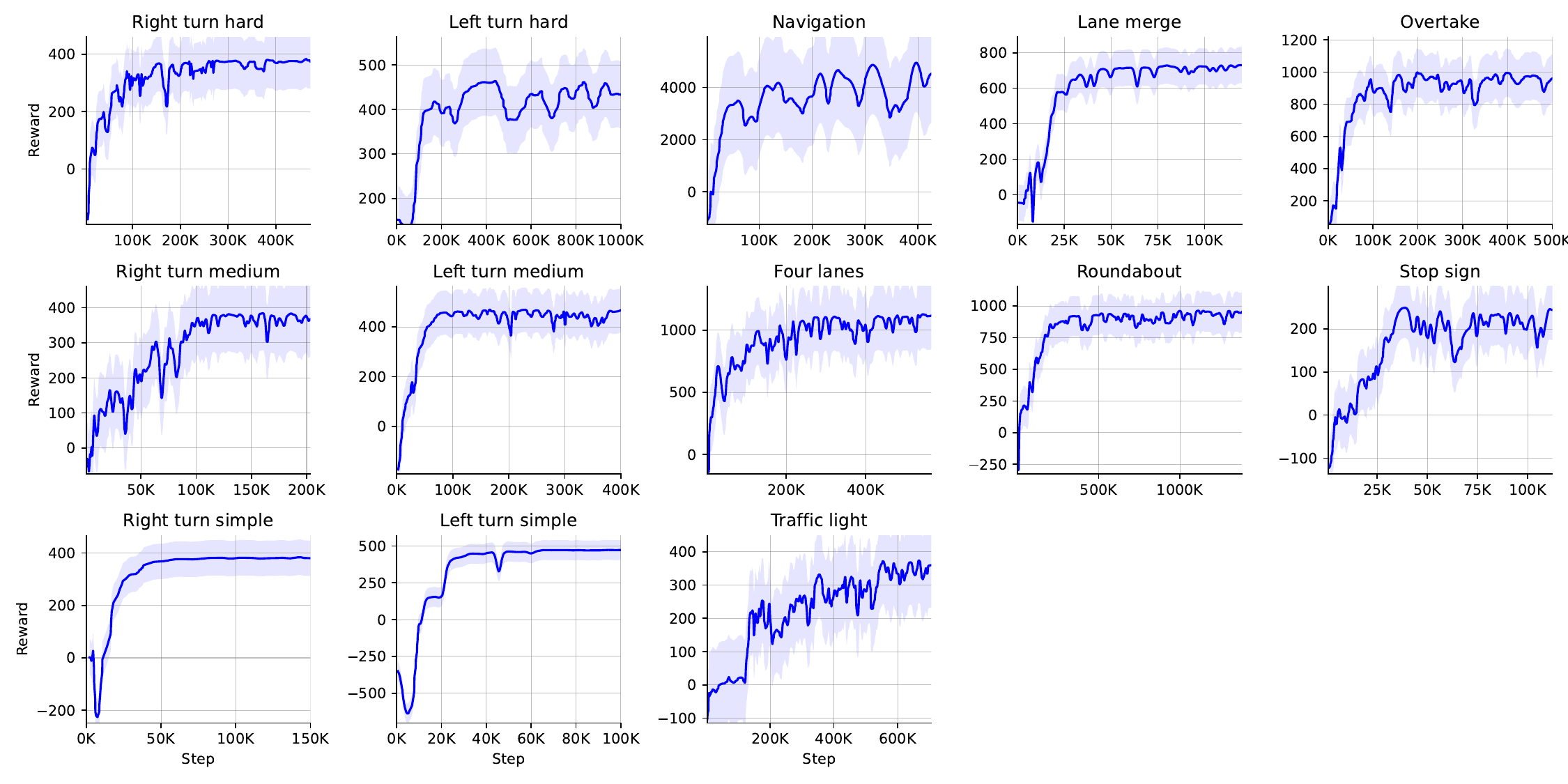}
    \caption{Reward curves of different tasks.}
    \label{fig:training reward}
\end{figure}

We use a small DreamerV3 model with only 18M parameters as the model backbone. A small DreamerV3 has 32 CNN multipliers, 512 GRU and MLP units, and the MLP has only two layers within its RSSM~\cite{hafner2023mastering}. The small memory overhead is around 22GB which allows us to train on a single NVIDIA 4090 GPU alongside running CARLA simulator. 

We train the agent on each task. The change in reward curves with respect to time steps is shown in \Cref{fig:training reward}. All the tasks here are trained with intention sharing and full observability. Simpler tasks with less traffic, such as `right turn simple' and `lane merge', typically converge within 50k steps (about 1 hour), whereas tasks involving denser, aggressive traffic flows, which require collision avoidance, take approximately 150k-200k steps to converge (about 3 to 4 hours). 

In our evaluation, we employ several metrics to rigorously assess the performance of autonomous driving agents executed within the CarDreamer tasks, detailed in \Cref{tab:metrics}. These metrics include:

\begin{itemize}
    \item \textbf{Success Rate:} This metric measures the percentage of episodes in which the ego vehicle successfully completes the task by reaching a destination point or traveling a predetermined distance without incident or out of lanes.
    
    % \item \textbf{Average Distance (m):} Represents the average distance traveled by the ego vehicle across all episodes before the episode terminates, either through task completion or due to a failure such as a collision or timeout.
    
    \item \textbf{Collision Rate (\%):} Calculates the percentage of episodes where the ego vehicle is involved in a collision.
    
    \item \textbf{Average Speed (m/s):} Measures the average speed maintained by the ego vehicle throughout the task. This metric is indicative of how efficiently the vehicle navigates the environment, balancing speed with safety.
    
    % \item \textbf{Waypoint Distance:} This metric quantifies the average divergence from the desired route waypoints. It assesses the vehicle's ability to adhere to the planned path, reflecting its navigation accuracy and precision in following the given trajectory.
\end{itemize}

It is worth noting that several tasks, such as `right turn' and `left turn', are notably challenging in environments with background traffic, where traffic flows aggressively and always disregards traffic rules and signs. This behavior increases the potential for collisions with the ego vehicle. Consequently, the AV must accurately predict future maneuvering of other vehicles to successfully navigate through the task.

\begin{figure}
    \centering
    \includegraphics[width=\textwidth]{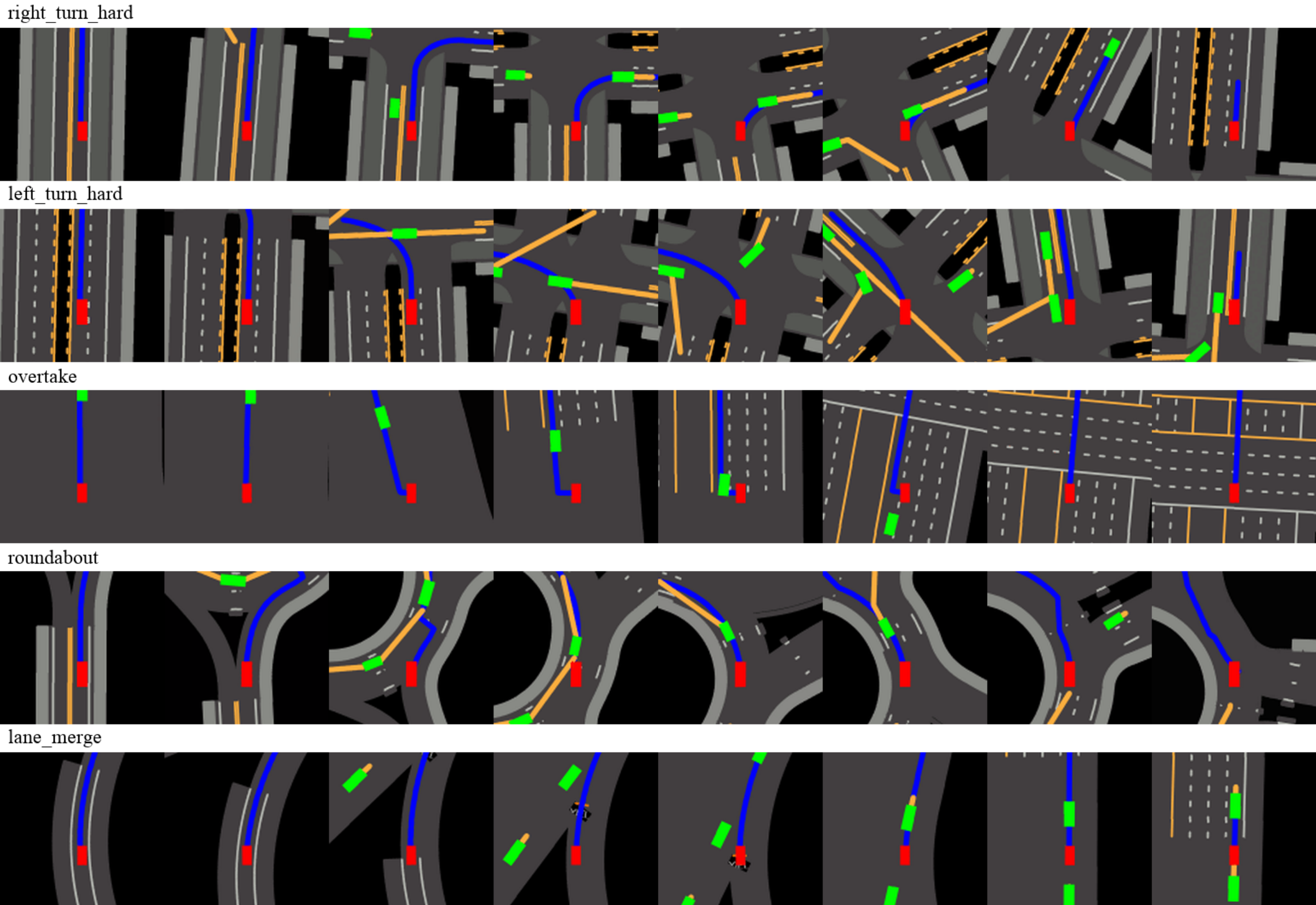}
    \caption{Sampled images during one episode in different tasks.}
    \label{fig:trained fig}
\end{figure}

    % \toprule
    %     Tasks & Success Rate & Avg. Distance (m) &  Collision Rate & Avg. Speed (m/s) & Wpt. Distance\\
    % \midrule
    %     Right turn hard & $ 97.63\%$  & 41.05 & 2.37 $\%$\ & 3.06 & 0.87 \\
    %     Right turn medium & $ 93.62\% $  & 40.82 & 6.37 $\%$\ & 3.25 & 0.85 \\
    %     Right turn simple & $ 100.00\% $  & 41.23 & 0.00 $\%$\ & 2.86 & 0.94 \\
    %     Left turn hard & $ 90.62\% $  & 46.51 & 9.38 $\%$\ & 1.49 & 0.86 \\
    %     Left turn medium & $ 84.85\% $  & 44.90 & 6.37 $\%$\ & 3.25 & 0.85 \\
    %     Left turn simple & $ 97.62\% $  & 45.23 & 0.00 $\%$\ & 2.00 & 1.28 \\
    %     Overtake & $ 93.02\% $  & 36.47 & 6.98 $\%$\ & 3.11 & 2.01 \\
    %     Four lane & $ 86.15\% $  & 94.97 & 12.31 $\%$\ & 3.13 & 0.91 \\
    %     Navigation & $ 91.46\% $  & 168.77 & 2.44 $\%$\ & 4.25 & 0.88 \\
    %     Lane merge & $ 89.38\% $  & 95.11 & 6.88 $\%$\ & 5.20 & 0.89 \\
    %     Roundabout & $ 84.16\% $  & 76.90 & 15.84 $\%$\ & 3.48 & 1.03 \\
    % \bottomrule

\begin{table*}
    \centering
    \caption{Performance metrics in different tasks.}
    \begin{threeparttable}
    \resizebox{0.8\textwidth}{!}{
    \begin{tabular}{cccccc} 
    \toprule
        Tasks & Success Rate  &  Collision Rate & Avg. Speed (m/s)\\
    \midrule
        Right turn simple &  99.21\% $\pm$ 0.49\%  & 0.00 \% $\pm$ 0.00\% & 3.19 $\pm$ 0.01 \\
        Right turn medium & 98.11 \% $\pm$ 0.22\%  & 1.89 \% $\pm$ 0.22\% & 2.99 $\pm$ 0.02 \\
        Right turn hard & 99.58 \% $\pm$ 0.42\%  & 0.42 \% $\pm$ 0.42\% & 2.92 $\pm$ 0.02 \\

        Left turn simple & 100.00\% $\pm$ 0.00\%  & 0.00\% $\pm$ 0.00\% & 3.21 $\pm$ 0.01 \\
        Left turn medium & 97.58\% $\pm$ 0.61\%  & 2.42\% $\pm$ 0.61\% & 3.04 $\pm$ 0.03 \\
        Left turn hard & 92.36\% $\pm$ 3.03\%  & 7.64 \% $\pm$ 3.03\% & 2.97 $\pm$ 0.01 \\

        Overtake & 100.00\% $\pm$ 0.00\%  & 0.00\% $\pm$ 0.00\% & 3.12 $\pm$ 0.03 \\
        Four lane & 96.83\% $\pm$ 3.17\%  & 3.17\% $\pm$ 3.17\% & 3.47 $\pm$ 0.00 \\
        Navigation & 80.95\% $\pm$ 6.41\%  & 7.58\% $\pm$ 4.01\% & 3.97 $\pm$ 0.08 \\
        Lane merge & 90.61\% $\pm$ 1.17\%  & 9.39\% $\pm$ 1.17\% & 5.17 $\pm$ 0.01 \\
        Roundabout & 96.00\% $\pm$ 2.31\%  & 4.00\% $\pm$ 2.31\% & 3.58 $\pm$ 0.00 \\
        Traffic lights & 92.65\% $\pm$ 0.85\%  & 0.00\% $\pm$ 0.00\% & 2.26 $\pm$ 0.03 \\
        Stop sign & 97.94\% $\pm$ 1.09\%  & 0.00\% $\pm$ 0.00\% & 1.80 $\pm$ 0.01 \\
    \bottomrule
    \end{tabular}
    }
    \end{threeparttable}
    \label{tab:metrics}
\end{table*}

\begin{figure}
    \centering

    % First image
    \begin{subfigure}[b]{\textwidth}
        \centering
        \includegraphics[width=\textwidth]{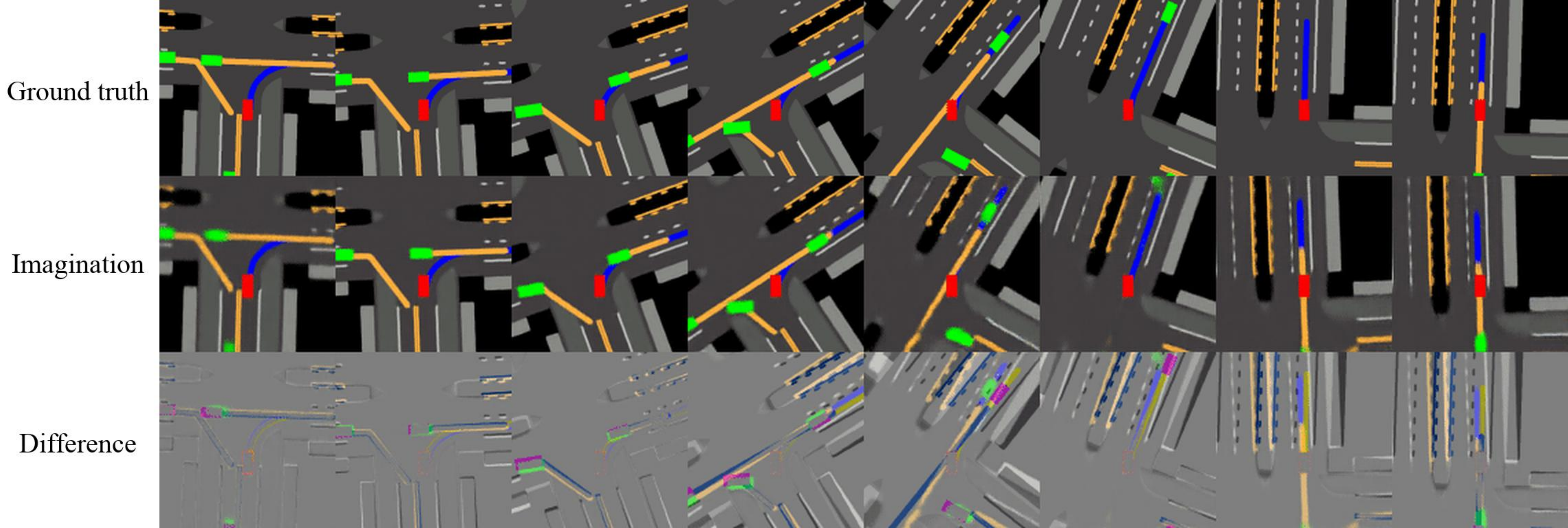}
        \caption{BEV}
    \end{subfigure}

    \vspace{1em} % Adjust vertical spacing between images as necessary

    % Second image
    \begin{subfigure}[b]{\textwidth}
        \centering
        \includegraphics[width=\textwidth]{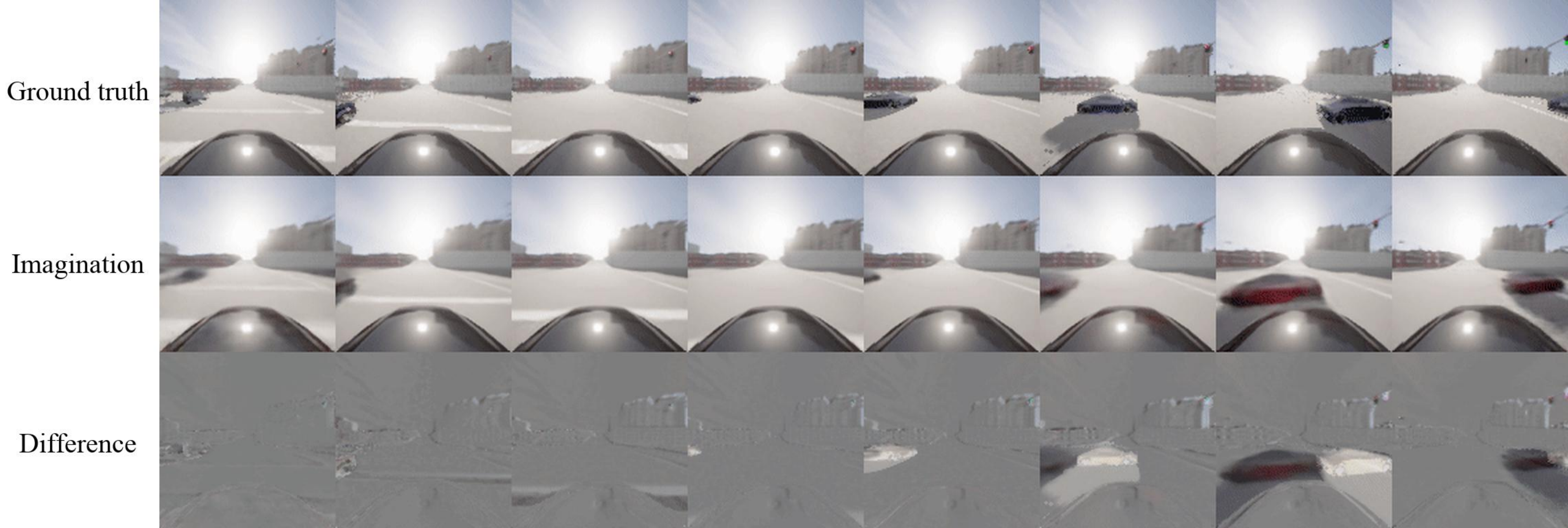}
        \caption{camera}
    \end{subfigure}

    \vspace{1em} % Adjust vertical spacing between images as necessary

    % Third image
    \begin{subfigure}[b]{\textwidth}
        \centering
        \includegraphics[width=\textwidth]{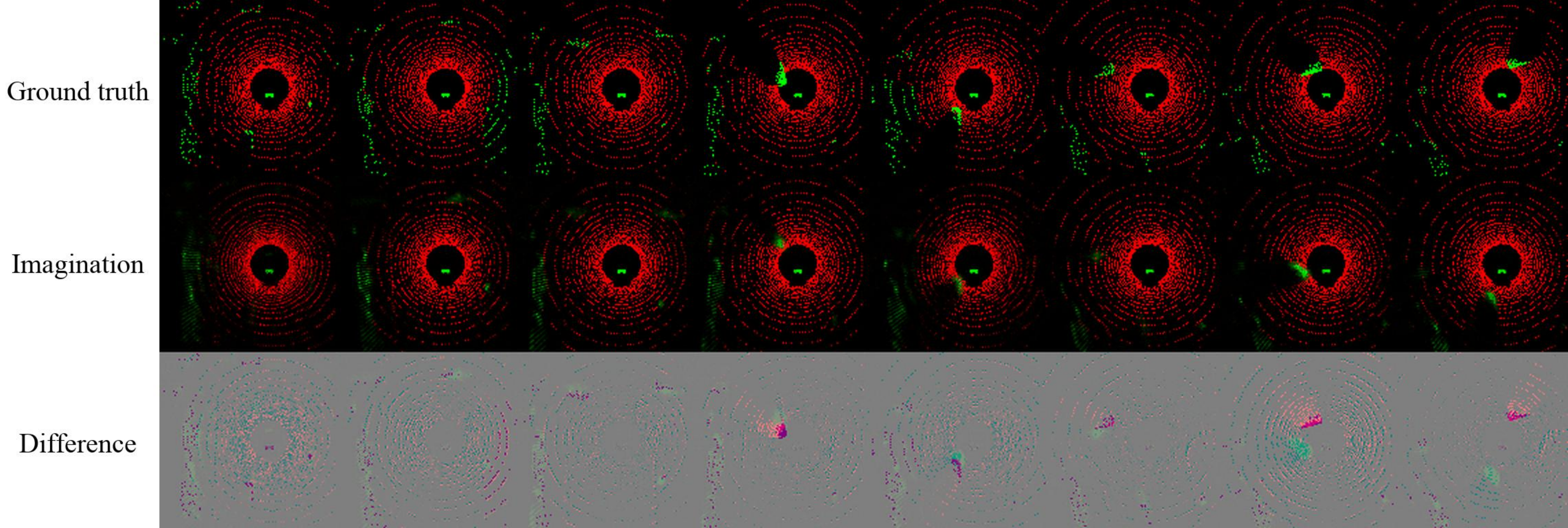}
        \caption{LiDAR}
    \end{subfigure}

    \caption{Comparison of the ground-truth observations and the ones imagined by WM in different modality settings}
    \label{fig:imagination}
\end{figure}
\subsection{Predictions in Different Observation Modalities}  \label{sub:open}

WM's imagination capability allows it to effectively predict future scenarios and manage potential events. To evaluate the WM's imagination performance with observations of different modalities, we conduct the experiments on the ``right turn hard'' task. We choose three different modalities: BEV, camera, and LiDAR. For each one, the WM is required to imagine the observations in a few future steps given the start state and a series of actions.

The results, illustrated in \Cref{fig:imagination}, compare the ground-truth images with the imagined ones across three modalities. The first row displays the ground truth observation images, the second row the WM's imagined outcomes, and the third row the differences between them. We selected frames within an imagination horizon of up to 64 time steps.

The findings demonstrate the WM's proficiency in accurately predicting the future despite the different modalities. In the BEV experiment (a), the WM precisely predicted the positions and trajectories of vehicles moving straight and making right turns, as well as the rotation and translation of the BEV with respect to the ego vehicle. Similarly, in camera and LiDAR settings, WM successfully predicts a vehicle driving in front of the ego vehicle.

\subsection{Benefits of V2V Communication}  \label{sub:com}

A distinctive feature of CarDreamer is its ability to facilitate easy customization of the level to which vehicles communicate. Vehicles can share FOV views, leading to different observability. Besides, they can even share intentions (represented by vehicles' planned waypoints) for better planning. We utilize this feature to evaluate the impact of communication. An agent is trained and tested on the ``right turn hard'' task under different settings, i.e., different observability and whether it has access to others' intentions. The ``right turn hard'' task is particularly suitable for testing observability and intention communication due to the dense traffic and frequent potential for collisions from vehicles outside the FOV.

\begin{figure}
    \centering
    \includegraphics[width=0.8\textwidth]{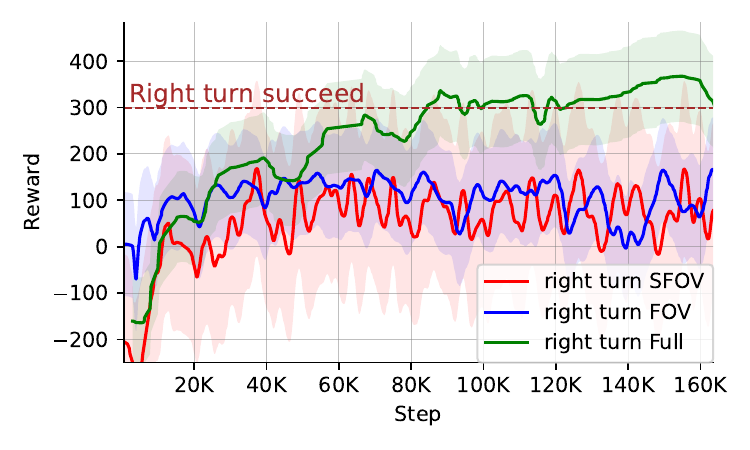}
    \caption{Reward curves in different observability settings.}
    \label{fig:reward for ob and wpt}
\end{figure}

\begin{figure}
    \centering
    \includegraphics[width=\textwidth]{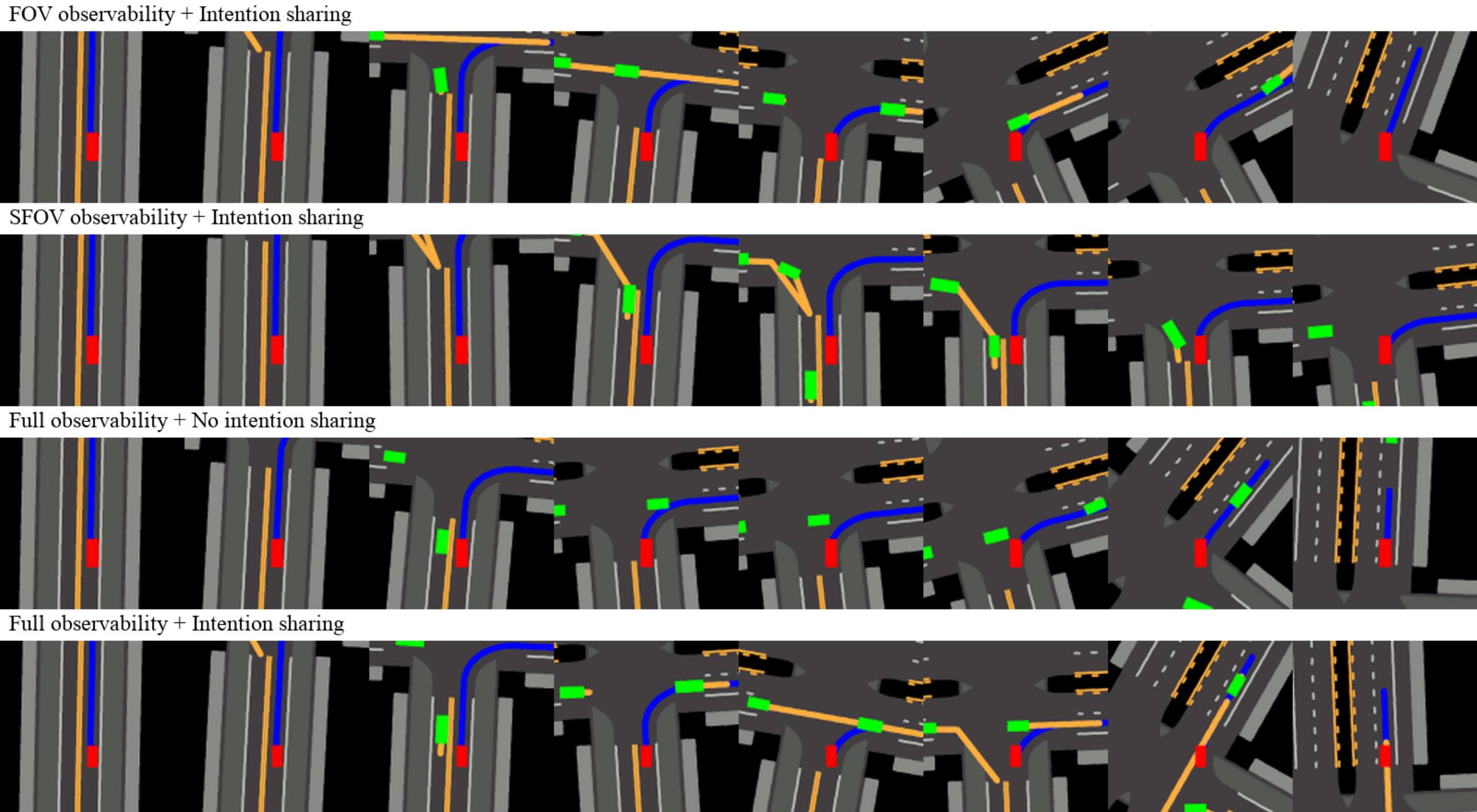}
    \caption{Sampled images during one episode in different communication and observability settings.}
    \label{fig:gif of observability}
\end{figure}

The reward curves are shown in \Cref{fig:reward for ob and wpt} and the performance metrics are shown in \Cref{tab:observability_ablation} and \Cref{tab:waypoint_ablation}. Note that the successful right turn is approximately indicated by the rewards exceeding 300. The results show that limited observability or lack of intention sharing impedes the agent from completing the task. The success rates of tasks with the above limits are significant lower than the task with full information. The evenly sampled images during one episode (shown in \Cref{fig:gif of observability}) provides a good explanation: the agent adopts a conservative and sub-optimal policy. For example, in the 'FOV' and 'SFOV' setting in \Cref{fig:gif of observability}, the agent hesitates to move on since they have insufficient information about coming vehicles. In the 'No intention sharing' setting, the agent cuts in at an inappropriate timing that leads to a collision without shared intentions. In contrast, the complete information enables the ego vehicle to successfully execute the right turn.

% \begin{figure}
% \centering
% \begin{subfigure}{0.45\linewidth}
%     \centering
%     \includegraphics[width=\linewidth]{figures/reward_for_observability.pdf}
%     \caption{Right turn hard with different observability settings.}
% \end{subfigure}\hfil
% \begin{subfigure}{0.45\linewidth}
%     \centering
%     \includegraphics[width=\linewidth]{figures/reward_for_waypoints.pdf}
%     \caption{Right turn hard with different intention sharing settings.}
%  \end{subfigure}
%  % \begin{subfigure}[t]{0.45\linewidth}
%  %    \centering
%  %    \includegraphics[width=\linewidth]{figures/reward for multi-modality.pdf}
%  %    \caption[b]{Training reward curves for multi-modalities.}
%  % \end{subfigure}
%  % \caption{Rewards from training in the observability and communication experiments.}
%  \caption{Reward curves in different communication settings.}
%  \label{fig:reward for ob and wpt}
% \end{figure}

\begin{table}
    \centering
    \caption{Metrics in different observability settings (with intention sharing).}
    \begin{threeparttable}
    \begin{tabular}{cccccc} 
    \toprule
        Settings & Success Rate  &  Collision Rate & Avg. Speed (m/s)\\
    \midrule
        Full Observability & 99.58 \% $\pm$ 0.42\%  & 0.42 \% $\pm$ 0.42\% & 2.92 $\pm$ 0.02 \\
        FOV Observability  & 12.44 \% $\pm$ 1.79\%  & 87.56 \% $\pm$ 1.79\% & 2.78 $\pm$ 0.01 \\
        SFOV Observability & 29.52 \% $\pm$ 3.46\%  & 70.48 \% $\pm$ 3.46\% & 2.86 $\pm$ 0.09 \\
    \bottomrule
    \end{tabular}
    \end{threeparttable}
    \label{tab:observability_ablation}
\end{table}

\begin{table}
    \centering
    \caption{Metrics in different intention sharing settings (with full observability).}
    \begin{threeparttable}
    \begin{tabular}{cccccc} 
    \toprule
        Settings & Success Rate  &  Collision Rate & Avg. Speed (m/s)\\
    \midrule
        Intention Sharing & 99.58 \% $\pm$ 0.42\%  & 0.42 \% $\pm$ 0.42\% & 2.92 $\pm$ 0.02 \\
        No Intention Sharing & 90.71 \% $\pm$ 0.61\%  & 8.94 \% $\pm$ 0.58\% & 3.08 $\pm$ 0.00 \\
    \bottomrule
    \end{tabular}
    \end{threeparttable}
    \label{tab:waypoint_ablation}
\end{table}

\section{Conclusion}
In this work, we introduced CarDreamer, an open-source learning platform tailored for the development and evaluation of WM based RL algorithms in autonomous driving. CarDreamer offers a comprehensive set of built-in tasks, a flexible task development suite, and an integrated world model backbone, all aimed at facilitating rapid prototyping of driving tasks and algorithm testing within this specialized domain. With its modular design and diverse task configurations, CarDreamer establishes itself as a flexible and challenging testbed for assessing the performance of WM based autonomous driving systems. The experiments we conduct using our platform gives a comprehensive evaluation of the performance of DreamerV3 different driving tasks. We emphasize its predictive accuracy across different observation modalities and the significant impact of communication on performance.

Looking to the future, a promising avenue for further development involves the integration of curriculum learning \cite{bengio2009curriculum} and continual learning \cite{de2021continual} strategies. These approaches aim to systematically enhance the learning process by gradually increasing task complexity or continuously integrating new knowledge without forgetting previously acquired information. Furthermore, exploring advanced techniques such as transfer learning \cite{weiss2016survey} and meta-learning \cite{hospedales2021meta} could significantly improve the platform’s capabilities for few-shot adaptation to new environments. This would further augment CarDreamer’s utility in developing more generalized and robust autonomous driving approaches.

\bibliographystyle{unsrt}  
\bibliography{references}  %%% Remove comment to use the external .bib file (using bibtex).

\end{document}